\pdfoutput=1

\documentclass[11pt]{article}

\usepackage{acl}

\usepackage{times}
\usepackage{latexsym}

\usepackage[T1]{fontenc}

\usepackage[utf8]{inputenc}

\usepackage{microtype}

\usepackage{booktabs}
\usepackage{graphicx}
\usepackage{multirow}
\usepackage{makecell}

\newcommand{\mypar}[1]{\vspace{0.2em}\noindent\textbf{#1}}

\title{Unsupervised Syntactically Controlled Paraphrase Generation with Abstract Meaning Representations}

\author{
  Kuan-Hao Huang\thanks{$\;$ The authors contribute equally.}$^{\;\;\dagger}$ \ \ \ Varun Iyer\footnotemark[1]$^{\;\;\diamond}$ \ \ \ Anoop Kumar$^{\ddagger}$ \\
  {\bf Sriram Venkatapathy$^{\ddagger}$} \ \ \ {\bf Kai-Wei Chang$^{\dagger\ddagger}$}\vspace{0.2em} \ \ \ {\bf Aram Galstyan$^{\ddagger}$}  \\
  $^{\dagger}$University of California, Los Angeles \\
  $^{\diamond}$Johns Hopkins University, $^{\ddagger}$Amazon Alexa AI\vspace{0.1em}\\
  \texttt{\{khhuang, kwchang\}@cs.ucla.edu}, \ \ \ \texttt{viyer3@jhu.edu} \\
  \texttt{\{anooamzn, vesriram, kaiwec, argalsty\}@amazon.com} \\
}

\begin{document}
\maketitle

\begin{abstract}
Syntactically controlled paraphrase generation has become an emerging research direction in recent years. Most existing approaches require annotated paraphrase pairs for training and are thus costly to extend to new domains. Unsupervised approaches, on the other hand, do not need paraphrase pairs but suffer from relatively poor performance in terms of syntactic control and quality of generated paraphrases. In this paper, we demonstrate that leveraging Abstract Meaning Representations (AMR) can greatly improve the performance of unsupervised syntactically controlled paraphrase generation.
Our proposed model, \textbf{AMR}-enhanced \textbf{P}araphrase \textbf{G}enerator (AMRPG), separately encodes the AMR graph and the constituency parse of the input sentence into two disentangled semantic and syntactic embeddings.
A decoder is then learned to reconstruct the input sentence from the semantic and syntactic embeddings. Our experiments show that AMRPG generates more accurate syntactically controlled paraphrases, both quantitatively and qualitatively, compared to the existing unsupervised approaches. 
We also demonstrate that the paraphrases generated by AMRPG can be used for data augmentation to improve the robustness of NLP models.
\end{abstract}
\section{Introduction}

\begin{figure}
    \centering
    \includegraphics[width=.48\textwidth]{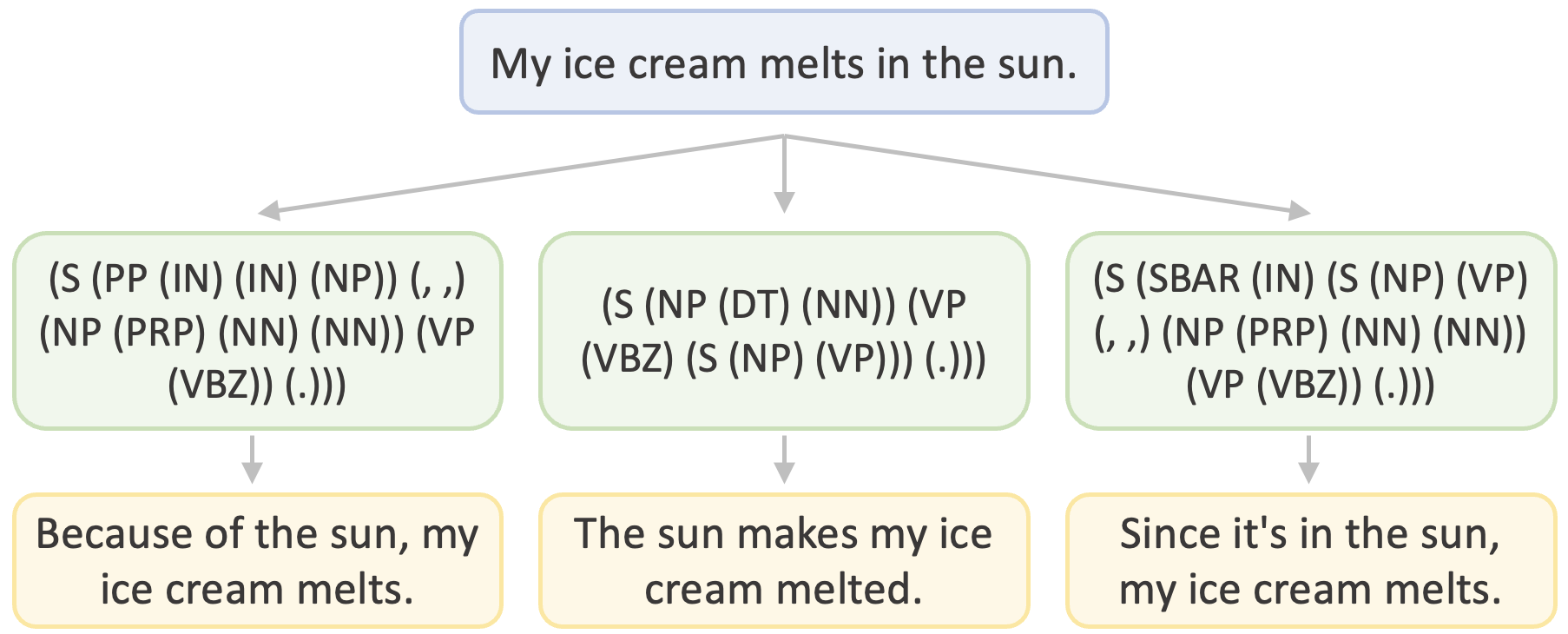}
    \caption{An illustration of syntactically controlled paraphrase generation. Given a source sentence and different parse specifications, the model generates different paraphrases following the parse specifications.}
    \label{fig:syn-demo}
    \vspace{-0.5em}
\end{figure}

Syntactically controlled paraphrase generation approaches aim to control the format of generated paraphrases by taking into account  additional parse specifications as the inputs, as illustrated by Figure~\ref{fig:syn-demo}.
It has attracted increasing attention in recent years since it can diversify the generated paraphrases and benefit a wide range of NLP applications \cite{Iyyer18scpn,Huang21synpg,Sun21aesop}, including task-oriented dialog generation \cite{Gao20dialoggen}, creative generation \cite{Tian21creativegen}, and model robustness \cite{Huang21synpg}.

Recent works have shown success in training syntactically controlled paraphrase generators \cite{Iyyer18scpn,Chen19cgen,Kumar20sgcp,Sun21aesop}.
Although their models can generate high-quality paraphrases and achieve good syntactic control ability, 
the training process needs a large amount of supervised data, e.g., parallel paraphrase pairs.
Annotating paraphrase pairs is usually expensive because it requires intensive domain knowledge and high-level semantic understanding.
Due to the difficulty in collecting parallel data, the ability of supervised approaches are limited, especially when adapting to new domains. 

To reduce the annotation demand, unsupervised approaches can train syntactically controlled paraphrase generators without the need for parallel pairs \cite{Zhang19sivae,Bao19sepvae,Huang21synpg}.
Most of them achieve syntactic control by learning disentangled embeddings for semantics and syntax separately \cite{Bao19sepvae,Huang21synpg}.
However, without parallel data, it is challenging to learn a good disentanglement and capture semantics well.
As we will show later (Section~\ref{sec:genexp}), unsupervised approaches can generate bad paraphrases by mistakenly swapping object and subject of a sentence.

\begin{figure}
    \centering
    \includegraphics[width=.42\textwidth]{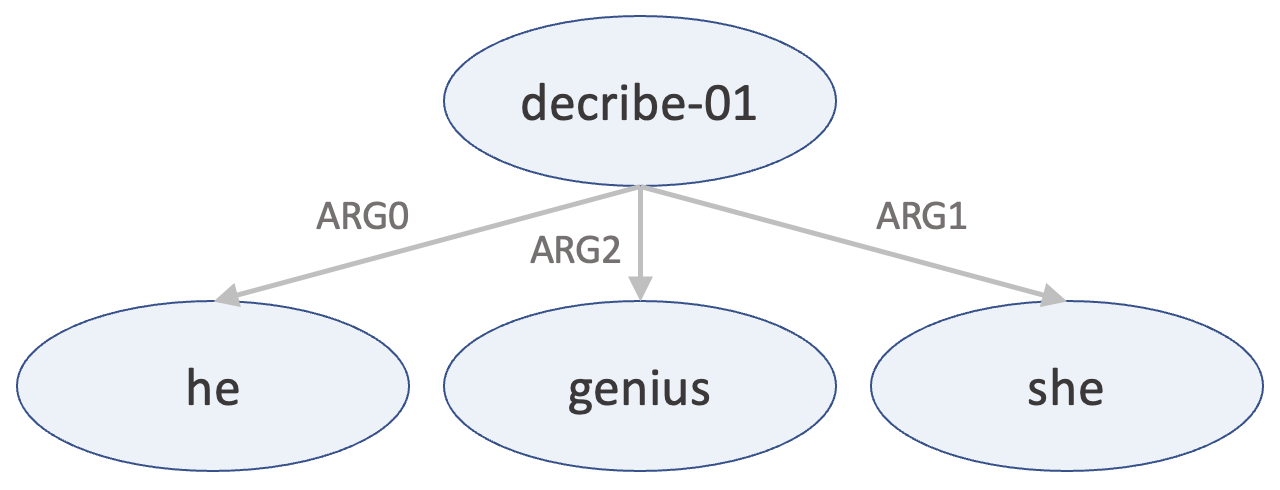}
    \caption{The same AMR graph for a pair of paraphrased sentences ``He described her as a genius.'' and ``She was a genius, according to his description.''}
    \label{fig:amr-demo}
    \vspace{-1.2em}
\end{figure}

In this work, we propose to use \emph{Abstract Meaning Representations (AMR) \cite{Banarescu13amr} to learn better disentangled semantic embeddings for unsupervised syntactically controlled paraphrase generation.}
AMR is a semantic graph structure that covers the abstract meaning of a sentence.
As shown in Figure~\ref{fig:amr-demo}, two sentences would have the same (or similar) AMR graph as long as they carry the same abstract meaning, even they are expressed with different syntactic structures. 
This property makes AMRs a good resource to capture sentence semantics.

Based on this, we design an \textbf{AMR}-enhanced \textbf{P}araphrase \textbf{G}enerator (AMRPG), which separately learns (1) \emph{semantic embeddings} with the AMR garphs extracted from the input sentence and (2) \emph{syntactic embeddings} from the constituency parse of the input sentence. Then, AMRPG trains a decoder to reconstruct the input sentence from the semantic and syntactic embeddings.
The reconstruction objective and the design of the disentanglement of semantics and the syntax makes AMRPG  learn to generate syntactically controlled paraphrases without using parallel pairs.
Our experiments show that AMRPG performs better syntactic control than existing unsupervised approaches.
Additionally, we demonstrate that the generated paraphrases of AMRPG can be used for data augmentation to improve the robustness of NLP models.
\looseness=-1
\section{Related Work}

\paragraph{Paraphrase generation.}
Traditional paraphrase generators are usually based on hand-crafted rules \cite{Barzilay03rule1} or seq2seq models \cite{Cao17seq2seq1,Gupta18seq2seq2,Fu19seq2seq3}.
To generate diverse paraphrases, different techniques are proposed, including random pattern embeddings \cite{Kumar19randompattern1}, latent space perturbation \cite{Roy19latent1,Zhang19sivae,Cao20latent2}, multi-round generation \cite{Lin21multiround}, reinforcement learning \cite{Liu20rl1}, prompt-tuning \cite{Chowdhury22prompt}, order control \cite{Goyal20order1}, and syntactic control \cite{Iyyer18scpn,Kumar20sgcp,Huang21synpg,Sun21aesop}.

\paragraph{Abstract meaning representation (AMR).}
Since AMR \cite{Banarescu13amr} captures high-level semantics, it has been applied for various NLP tasks, including summarization \cite{Sachan16amrsum}, dialogue modeling \cite{Bai20amrdm}, information extraction \cite{Zhang20amrie}.
Some works also focus on training high-quality AMR parsers with graph encoders \cite{Cai20amrparser2}, seq2seq models \cite{Konstas17amrparser4,Zhou20amrparser3}, and decoder-only models \cite{Bevilacqua21amrparse1}.

\section{Unsupervised Syntactically Controlled Paraphrase Generation}
\label{sec:method}

\subsection{Problem Formulation}

We follow previous works \cite{Iyyer18scpn,Huang21synpg} and consider constituency parses (without terminals) as the control signals.
Given a source sentence $s$ and a target parse $p$, the goal of the syntactically controlled paraphrase generator is to generate a target sentence $t$ which has similar semantics to the source sentence $s$ and has syntax following the parse~$p$.
In the unsupervised setting, the paraphrase generator cannot access any target sentences and target parses but only the source sentences and source parses during training.
\looseness=-1

\subsection{Proposed Method: AMRPG}
\begin{figure*}
    \centering
    \includegraphics[width=.96\textwidth]{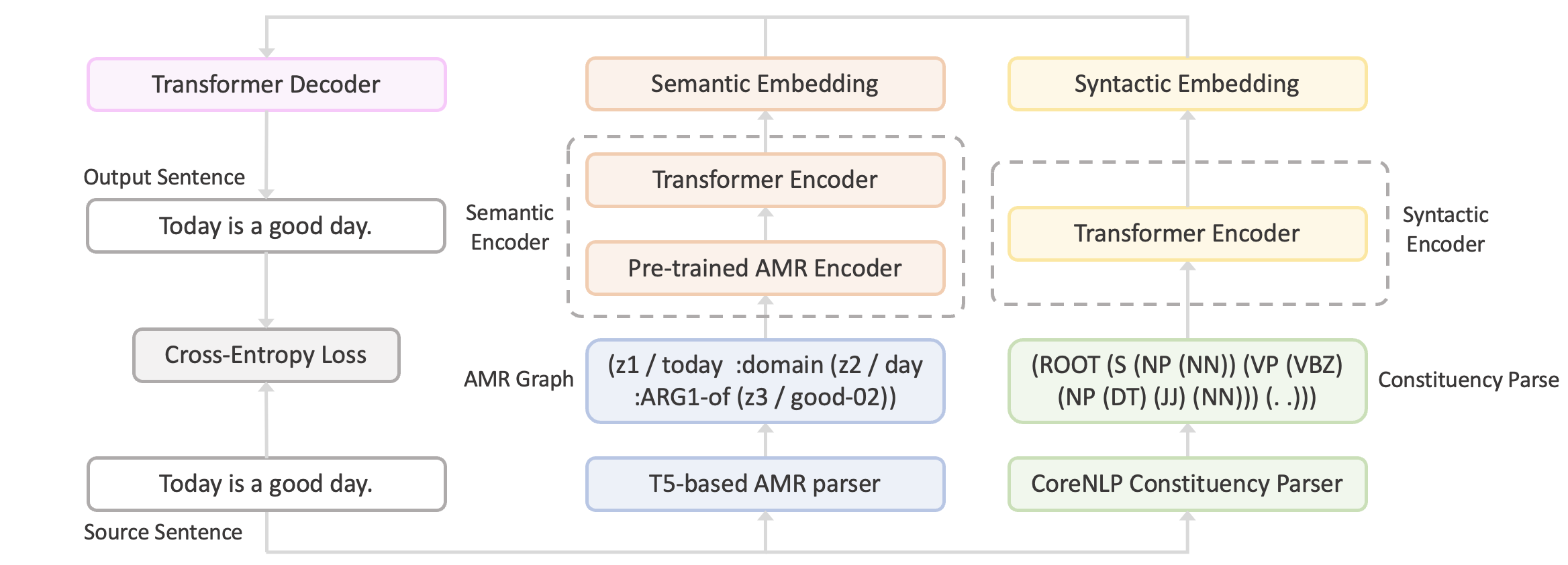}
    \caption{AMRPG's framwork. It separately encodes the AMR graph and the constituency parse of the input sentence into two disentangled semantic and syntactic embeddings.
A decoder is then learned to reconstruct the input sentence from the semantic and syntactic embeddings.}
    \label{fig:overview}
    \vspace{-1em}
\end{figure*}

Motivated by previous approaches \cite{Bao19sepvae,Huang21synpg}, we design AMRPG to learn separate embeddings for semantics and syntax, as illustrated by Figure~\ref{fig:overview}.
Then, AMRPG learns a decoder with the objective to reconstruct the source sentence. 
The challenge here is how to learn embeddings such that the semantic embedding contains only semantic information while the syntactic embedding contains only syntactic information.
We introduce the details as follows.

\paragraph{Semantic embedding.}
Given a source sentence, we first use a pre-trained AMR parser\footnote{{https://github.com/bjascob/amrlib-models}} to get its AMR graph.
Next, we use a semantic encoder to encode the AMR graph into the semantic embedding $e_{sem}$.
Specifically, the semantic encoder consists of two parts: a fixed pre-trained AMR encoder \cite{Ribeiro21amrenc} followed by a learnable Transformer encoder.
We additionally perform \emph{node masking} when training the semantic encoder.
Specifically, every node in the AMR graph has a probability to be masked out during training.
This can improve the robustness of AMRPG.

As mentioned above, two semantically similar sentences would have similar AMR graphs regardless of their syntax. This property encourages AMRPG to capture only semantic information in semantic embeddings.
Compared with previous work \cite{Huang21synpg}, which uses bag-of-words to learn the semantic embeddings, using AMR can capture semantics better and lead to better performance, as shown in Section~\ref{sec:exp}.

\paragraph{Syntactic embedding.}
Given a source sentence, we use the Stanford CoreNLP toolkit \cite{Manning14corenlp} to get its constituency parse.
Then, we remove all the terminals in the parse and learns a Transformer encoder to encode the parse into the syntactic embedding $e_{syn}$.
Since we remove the terminals, the syntactic embedding contains only the syntactic information of the source sentence.

\paragraph{Decoder.}
We train a Transformer decoder that takes the semantic embedding $e_{sem}$ and the syntactic embedding $e_{syn}$ as the input, and reconstructs the source sentence with a cross-entropy loss.
The reconstruction objective makes AMRPG not require parallel paraphrase pairs for training.

\paragraph{Inference.}
Given a source sentence $s$ and a target parse $p$, we use the semantic encoder to encode the AMR graph of $s$ into the semantic embedding, use the syntactic encoder to encode $p$ into the syntactic embedding, and use the decoder to generate the target sentence $t$.
\section{Experiments}
\label{sec:exp}

\subsection{Syntactically Controlled Paraphrase Generation}
\label{sec:genexp}

\paragraph{Datasets.}
We consider ParaNMT \cite{Gimpel18paranmt} for training and testing.
We use \emph{only the source sentences} in ParaNMT to train AMRPG and other unsupervised baselines, and use both the source sentences and target sentences to train supervised baselines.
To further test the model's ability to generalize to new domains, we directly use the models trained with ParaNMT to test on Quora \cite{Iyer17quora}, MRPC \cite{Dolan04mrpc}, and PAN \cite{Madnani12pan}

\paragraph{Evaluation metrics.}
Following the previous work \cite{Huang21synpg}, we consider the BLEU score to measure the similarity between the gold target sentences and the predicted target sentences, and consider the \emph{template matching accuracy}\footnote{Template matching accuracy is defined as the exact matching accuracy of top-2 levels of parse trees.} (TMA) to evaluate the goodness of syntactic control.
More details about the evaluation can be found in Appendix~\ref{app:eval}.

\paragraph{Baselines.}
We consider the following unsupervised models:  SIVAE \cite{Zhang19sivae}, SynPG \cite{Huang21synpg}, AMRPG, and T5-Baseline, which replaces the AMR encoder with a T5-encoder. We also consider SCPN \cite{Iyyer18scpn} as the supervised baseline.

\begin{table*}
\centering
\small
\aboverulesep = 0.6mm
\belowrulesep = 0.6mm
\setlength{\tabcolsep}{4.5pt}
\resizebox{0.88\textwidth}{!}{
\begin{tabular}{lcccccccc}
    \toprule
    \multirow{2}{*}{Model} 
    & \multicolumn{2}{c}{ParaNMT}
    & \multicolumn{2}{c}{Quora}
    & \multicolumn{2}{c}{PAN}
    & \multicolumn{2}{c}{MRPC} \\
    \cmidrule{2-9}
    & TMA & BLEU & TMA & BLEU & TMA & BLEU & TMA & BLEU \\
    \midrule
    \multicolumn{9}{c}{\emph{Unsupervised Approaches (without using parallel pairs)}} \\
    \midrule
    SIVAE \cite{Zhang19sivae}
    & 30.0 & 12.8 & 48.3 & 13.1 & 26.6 & 11.8 & 21.5 &  5.1 \\
    SynPG \cite{Huang21synpg} 
    & 71.0 & 32.2 & 82.6 & 33.2 & \textbf{66.3} & 26.4 & \textbf{74.0} & 26.2 \\
    T5-Baseline
    & 57.1 & 22.8 & 66.1 & 22.2 & 55.3 & 21.0 & 66.2 & 18.8 \\
    AMRPG
    & \textbf{74.3} & \textbf{39.1} & \textbf{84.8} & \textbf{33.9} & 65.6 & \textbf{31.0} & 71.9 & \textbf{34.8} \\
    
    \midrule
    \multicolumn{9}{c}{\emph{Unsupervised Approaches (using target domain source sentences)}} \\
    \midrule
    SynPG \cite{Huang21synpg} 
    & -    & -    & 86.3 & 44.4 & 66.4 & 34.2 & \textbf{80.7} & 44.6 \\
    AMRPG
    & -    & -    & \textbf{86.5} & \textbf{45.4} & \textbf{67.5} & \textbf{37.6} & 76.8 & \textbf{45.9} \\
    
    \midrule
    \multicolumn{9}{c}{\emph{Supervised Approaches (using additional parallel pairs in ParaNMT; not compariable to ours)}} \\
    \midrule
    SCPN \cite{Iyyer18scpn}
    & 83.9 & 58.3 & 87.1 & 41.0 & 72.3 & 37.6 & 80.1 & 41.8 \\
    
    \bottomrule
\end{tabular}}
\caption{Results of syntactically controlled paraphrase generation. AMRPG performs the best among all unsupervised approaches and can outperform supervised approaches when considering the target domain source sentences.}
\label{tab:paragen}
\end{table*}

\begin{table*}[t!]
\scriptsize
\centering
\aboverulesep = 0.2mm
\belowrulesep = 0.5mm
\setlength{\tabcolsep}{7pt}
\resizebox{.85\textwidth}{!}{
\begin{tabular}{ll}
    \toprule
    \textbf{Input} & The dog chased the cat on the street. \\
    \textbf{Parse template} & \texttt{(S(NP(DT)(NN))(VP(VBN)(PP))(.))} \\
    \textbf{Target} & The cat was chased by the dog on the street. \\
    \textbf{SynPG} & The dog was chased by the cat on the street. \\
    \textbf{AMRPG} & The cat was chased by a dog in the street. \\
    \midrule
    \textbf{Input} & John will send a gift to Tom when Christmas comes. \\
    \textbf{Parse template} & \texttt{(S(SBAR (WHADVP)(S))(,)(NP(NNP))(VP(MD)(VP))(.))} \\ 
    \textbf{Target} & When Christmas comes, John will send a gift to Tom. \\
    \textbf{SynPG} & When Tom comes, John will send a gift to Christmas. \\
    \textbf{AMRPG} & When Christmas comes, John will send a gift to Tom. \\
    \bottomrule
\end{tabular}}
\caption{Paraphrase examples generated by SynPG and AMRPG. AMRPG captures semantics better and generates higher quality of paraphrases than SynPG.}
\label{tab:example}
\vspace{-1em}
\end{table*}


\paragraph{Results.}
Table~\ref{tab:paragen} shows the results of syntactically controlled paraphrase generation.
AMRPG performs the best among the unsupervised approaches.
Specifically, AMRPG outperforms SynPG, the state-of-the-art unsupervised model,  with a large gap in terms of BLEU score.
This justifies that using AMR can learn better disentangled embeddings and capture semantics better.

We observe that there is indeed a performance gap between AMRPG and SCPN (supervised baseline).
However, since AMRPG is an unsupervised model, it is possible to use the source sentences from the target domains to further fine-tune AMRPG without additional annotation cost. 
As shown in the table, AMRPG with further fine-tuning can achieve even better performance than SCPN when considering domain adaptation (Quora, MRPC, and PAN).
This demonstrates the flexibility and the potential of unsupervised paraphrase models.

\mypar{Qualitative examples.}
Table~\ref{tab:example} lists some paraphrases generated by SynPG and AMRPG.
As we mentioned in Section~\ref{sec:method}, SynPG uses bag-of-words to learn semantic embeddings and therefore SynPG is easy to get confused about the relations between entities or mistake the subject for the object.
In contrast, AMRPG can preserve more semantics.

\subsection{Improving Robustness of NLP Models}

We demonstrate that the paraphrases generated by AMRPG can improve the robustness of NLP models by data augmentation.
Following the setting of previous work \cite{Huang21synpg}, we consider three classification tasks in GLUE \cite{Wang19glue}: MRPC, RTE, and SST-2. 
We compare three baselines: (1) the classifier trained with original training data, (2) the classifier trained with original training data and augmented data generated by SynPG, and (3) the classifier trained with original training data and augmented data generated by AMRPG.
Specifically, for every instance in the original training data, we generate four paraphrases as the augmented examples by considering four common syntactic templates.
More details can be found in Appendix~\ref{app:robtrain}.

Table~\ref{tab:robustness} shows the clean accuracy and the broken rate (the percentage of examples being  attacked) after attacked by the syntactically adversarial examples\footnote{Appendix \ref{app:adv} has more details of the adversarial attack.} generated with SCPN \cite{Iyyer18scpn}. 
Although the classifiers trained with data augmentation have slightly worse clean accuracy, they have significantly lower broken rates, which implies that data augmentation improves the model robustness.
Also, data augmentation with AMRPG performs better than data augmentation with SynPG in terms of the broken rate.
We attribute this to the better quality of paraphrase generation of AMRPG.

\begin{table}
\centering
\small
\aboverulesep = 0.6mm
\belowrulesep = 0.6mm
\setlength{\tabcolsep}{3pt}
\resizebox{0.44\textwidth}{!}{
\begin{tabular}{lcccccc}
    \toprule
    \multirow{2}{*}{Model} 
    & \multicolumn{2}{c}{MRPC}
    & \multicolumn{2}{c}{RTE} 
    & \multicolumn{2}{c}{SST-2} \\
    \cmidrule{2-7}
    & Acc. & Brok. & Acc. & Brok. & Acc. & Brok. \\
    \midrule
    Base    & \textbf{83.3} & 52.9 & \textbf{62.1} & 58.1 & \textbf{92.2} & 38.8 \\
    \midrule
    + SynPG & 80.6 & 42.2 & 61.7 & 40.3 & 91.5 & 38.5 \\
    + AMRPG & 80.6 & \textbf{38.3} & 58.8 & \textbf{39.3} & 91.6 & \textbf{36.7} \\
    \bottomrule
\end{tabular}}
\caption{Augmenting paraphrases generated by AMRPG improves the robustness of NLP models. Acc denotes the clean 
accuracy (the higher is the better).
Brok denotes the percentage of examples being successfully attacked (the lower is the better).}
\vspace{-1.3em}
\label{tab:robustness}
\end{table}
\section{Conclusion}

We propose AMRPG that utilizes AMR to learn a better disentanglement of semantics and syntax without using any parallel data.
This enables AMRPG to captures semantics better and generate more accurate syntactically controlled paraphrases than existing unsupervised approaches. 
We also demonstrate that how to apply AMRPG to improve the robustness of NLP models.
\section*{Limitations}

Our goal is to demonstrate the potential of AMR for syntactically controlled paraphrase generation. The current experimental setting follows previous works \cite{Iyyer18scpn,Huang21synpg}, which considers the \emph{full} constituency parses as the control signals.
In real applications, getting full constituency parses before the paraphrase generation process might take additional efforts.
One potential solution is to consider relatively noisy or simplified parse specifications \cite{Sun21aesop}.
In addition, some parse specifications can be inappropriate for certain source sentences (e.g., the source sentence is long but the target parse is short).
How to score and reject some of the given parse specifications is still an open research question.
Finally, although training AMRPG does not require any parallel paraphrase pairs, it does require a pre-trained AMR parser, which can be a potential cost for training AMRPG.

\section*{Broader Impacts}
Our proposed method focuses on improving syntactically controlled paraphrase generation. It is intended to be used to improve the robustness of models and facilitate language generation for applications with positive social impacts. All the experiments are conducted on open benchmark datasets. However, it is known that the models trained with a large text corpus could capture the bias reflecting the training data. It is possible for our model to potentially generate offensive or biased content learned from the data. We suggest to carefully examining the potential bias before deploying models in any real-world applications. 


\bibliography{custom}

\begin{thebibliography}{35}
\expandafter\ifx\csname natexlab\endcsname\relax\def\natexlab#1{#1}\fi

\bibitem[{Bai et~al.(2021)Bai, Chen, Song, and Zhang}]{Bai20amrdm}
Xuefeng Bai, Yulong Chen, Linfeng Song, and Yue Zhang. 2021.
\newblock Semantic representation for dialogue modeling.
\newblock In \emph{Proceedings of the 59th Annual Meeting of the Association
  for Computational Linguistics and the 11th International Joint Conference on
  Natural Language Processing (ACL/IJCNLP)}.

\bibitem[{Banarescu et~al.(2013)Banarescu, Bonial, Cai, Georgescu, Griffitt,
  Hermjakob, Knight, Koehn, Palmer, and Schneider}]{Banarescu13amr}
Laura Banarescu, Claire Bonial, Shu Cai, Madalina Georgescu, Kira Griffitt, Ulf
  Hermjakob, Kevin Knight, Philipp Koehn, Martha Palmer, and Nathan Schneider.
  2013.
\newblock Abstract meaning representation for sembanking.
\newblock In \emph{Proceedings of the 7th Linguistic Annotation Workshop and
  Interoperability with Discourse (LAW-ID@ACL)}.

\bibitem[{Bao et~al.(2019)Bao, Zhou, Huang, Li, Mou, Vechtomova, Dai, and
  Chen}]{Bao19sepvae}
Yu~Bao, Hao Zhou, Shujian Huang, Lei Li, Lili Mou, Olga Vechtomova, Xin{-}Yu
  Dai, and Jiajun Chen. 2019.
\newblock Generating sentences from disentangled syntactic and semantic spaces.
\newblock In \emph{Proceedings of the 57th Conference of the Association for
  Computational Linguistics (ACL)}.

\bibitem[{Barzilay and Lee(2003)}]{Barzilay03rule1}
Regina Barzilay and Lillian Lee. 2003.
\newblock Learning to paraphrase: An unsupervised approach using
  multiple-sequence alignment.
\newblock In \emph{Proceedings of the 2003 Conference of the North American
  Chapter of the Association for Computational Linguistics: Human Language
  Technologies (NAACL-HLT)}.

\bibitem[{Bevilacqua et~al.(2021)Bevilacqua, Blloshmi, and
  Navigli}]{Bevilacqua21amrparse1}
Michele Bevilacqua, Rexhina Blloshmi, and Roberto Navigli. 2021.
\newblock One {SPRING} to rule them both: Symmetric {AMR} semantic parsing and
  generation without a complex pipeline.
\newblock In \emph{Thirty-Fifth {AAAI} Conference on Artificial Intelligence
  (AAAI)}.

\bibitem[{Cai and Lam(2020)}]{Cai20amrparser2}
Deng Cai and Wai Lam. 2020.
\newblock {AMR} parsing via graph-sequence iterative inference.
\newblock In \emph{Proceedings of the 58th Annual Meeting of the Association
  for Computational Linguistics (ACL)}.

\bibitem[{Cao and Wan(2020)}]{Cao20latent2}
Yue Cao and Xiaojun Wan. 2020.
\newblock Divgan: Towards diverse paraphrase generation via diversified
  generative adversarial network.
\newblock In \emph{Findings of the Association for Computational Linguistics:
  (EMNLP-Findings)}.

\bibitem[{Cao et~al.(2017)Cao, Luo, Li, and Li}]{Cao17seq2seq1}
Ziqiang Cao, Chuwei Luo, Wenjie Li, and Sujian Li. 2017.
\newblock Joint copying and restricted generation for paraphrase.
\newblock In \emph{Proceedings of the Thirty-First {AAAI} Conference on
  Artificial Intelligence (AAAI)}.

\bibitem[{Chen et~al.(2019)Chen, Tang, Wiseman, and Gimpel}]{Chen19cgen}
Mingda Chen, Qingming Tang, Sam Wiseman, and Kevin Gimpel. 2019.
\newblock Controllable paraphrase generation with a syntactic exemplar.
\newblock In \emph{Proceedings of the 57th Conference of the Association for
  Computational Linguistics (ACL)}.

\bibitem[{Chowdhury et~al.(2022)Chowdhury, Zhuang, and
  Wang}]{Chowdhury22prompt}
Jishnu~Ray Chowdhury, Yong Zhuang, and Shuyi Wang. 2022.
\newblock Novelty controlled paraphrase generation with retrieval augmented
  conditional prompt tuning.
\newblock In \emph{Proceedings of the Thirty-Sixth {AAAI} Conference on
  Artificial Intelligence (AAAI)}.

\bibitem[{Dolan et~al.(2004)Dolan, Quirk, and Brockett}]{Dolan04mrpc}
Bill Dolan, Chris Quirk, and Chris Brockett. 2004.
\newblock Unsupervised construction of large paraphrase corpora: Exploiting
  massively parallel news sources.
\newblock In \emph{20th International Conference on Computational Linguistics
  (COLING)}.

\bibitem[{Fu et~al.(2019)Fu, Feng, and Cunningham}]{Fu19seq2seq3}
Yao Fu, Yansong Feng, and John~P. Cunningham. 2019.
\newblock Paraphrase generation with latent bag of words.
\newblock In \emph{Advances in Neural Information Processing Systems 32: Annual
  Conference on Neural Information Processing Systems 2019 (NeurIPS)}.

\bibitem[{Gao et~al.(2020)Gao, Zhang, Ou, and Yu}]{Gao20dialoggen}
Silin Gao, Yichi Zhang, Zhijian Ou, and Zhou Yu. 2020.
\newblock Paraphrase augmented task-oriented dialog generation.
\newblock In \emph{Proceedings of the 58th Annual Meeting of the Association
  for Computational Linguistics (ACL)}.

\bibitem[{Goyal and Durrett(2020)}]{Goyal20order1}
Tanya Goyal and Greg Durrett. 2020.
\newblock Neural syntactic preordering for controlled paraphrase generation.
\newblock In \emph{Proceedings of the 58th Annual Meeting of the Association
  for Computational Linguistics (ACL)}.

\bibitem[{Gupta et~al.(2018)Gupta, Agarwal, Singh, and Rai}]{Gupta18seq2seq2}
Ankush Gupta, Arvind Agarwal, Prawaan Singh, and Piyush Rai. 2018.
\newblock A deep generative framework for paraphrase generation.
\newblock In \emph{Proceedings of the Thirty-Second {AAAI} Conference on
  Artificial Intelligence (AAAI)}.

\bibitem[{Huang and Chang(2021)}]{Huang21synpg}
Kuan{-}Hao Huang and Kai{-}Wei Chang. 2021.
\newblock Generating syntactically controlled paraphrases without using
  annotated parallel pairs.
\newblock In \emph{Proceedings of the 16th Conference of the European Chapter
  of the Association for Computational Linguistics (EACL)}.

\bibitem[{Iyer et~al.(2017)Iyer, Dandekar, and el~Csernai}]{Iyer17quora}
Shankar Iyer, Nikhil Dandekar, and Korn el~Csernai. 2017.
\newblock First quora dataset release: Question pairs.
\newblock \emph{data.quora.com}.

\bibitem[{Iyyer et~al.(2018)Iyyer, Wieting, Gimpel, and
  Zettlemoyer}]{Iyyer18scpn}
Mohit Iyyer, John Wieting, Kevin Gimpel, and Luke Zettlemoyer. 2018.
\newblock Adversarial example generation with syntactically controlled
  paraphrase networks.
\newblock In \emph{Proceedings of the 2018 Conference of the North American
  Chapter of the Association for Computational Linguistics: Human Language
  Technologies (NAACL-HLT)}.

\bibitem[{Konstas et~al.(2017)Konstas, Iyer, Yatskar, Choi, and
  Zettlemoyer}]{Konstas17amrparser4}
Ioannis Konstas, Srinivasan Iyer, Mark Yatskar, Yejin Choi, and Luke
  Zettlemoyer. 2017.
\newblock Neural {AMR:} sequence-to-sequence models for parsing and generation.
\newblock In \emph{Proceedings of the 55th Annual Meeting of the Association
  for Computational Linguistics (ACL)}.

\bibitem[{Kumar et~al.(2020)Kumar, Ahuja, Vadapalli, and
  Talukdar}]{Kumar20sgcp}
Ashutosh Kumar, Kabir Ahuja, Raghuram Vadapalli, and Partha~P. Talukdar. 2020.
\newblock Syntax-guided controlled generation of paraphrases.
\newblock \emph{Transactions of the Association for Computational Linguistics},
  8:330--345.

\bibitem[{Kumar et~al.(2019)Kumar, Bhattamishra, Bhandari, and
  Talukdar}]{Kumar19randompattern1}
Ashutosh Kumar, Satwik Bhattamishra, Manik Bhandari, and Partha~P. Talukdar.
  2019.
\newblock Submodular optimization-based diverse paraphrasing and its
  effectiveness in data augmentation.
\newblock In \emph{Proceedings of the 2019 Conference of the North American
  Chapter of the Association for Computational Linguistics: Human Language
  Technologies (NAACL-HLT)}.

\bibitem[{Lin and Wan(2021)}]{Lin21multiround}
Zhe Lin and Xiaojun Wan. 2021.
\newblock Pushing paraphrase away from original sentence: {A} multi-round
  paraphrase generation approach.
\newblock In \emph{Findings of the Association for Computational Linguistics
  (ACL/IJCNLP-Findings)}.

\bibitem[{Liu et~al.(2020)Liu, Yang, Xiong, Zhang, Meng, Hu, Xu, and
  Chen}]{Liu20rl1}
Mingtong Liu, Erguang Yang, Deyi Xiong, Yujie Zhang, Yao Meng, Changjian Hu,
  Jinan Xu, and Yufeng Chen. 2020.
\newblock A learning-exploring method to generate diverse paraphrases with
  multi-objective deep reinforcement learning.
\newblock In \emph{Proceedings of the 28th International Conference on
  Computational Linguistics (COLING)}.

\bibitem[{Madnani et~al.(2012)Madnani, Tetreault, and Chodorow}]{Madnani12pan}
Nitin Madnani, Joel~R. Tetreault, and Martin Chodorow. 2012.
\newblock Re-examining machine translation metrics for paraphrase
  identification.
\newblock In \emph{Proceedings of the 2012 Conference of the North American
  Chapter of the Association for Computational Linguistics: Human Language
  Technologies (NAACL-HLT)}.

\bibitem[{Manning et~al.(2014)Manning, Surdeanu, Bauer, Finkel, Bethard, and
  McClosky}]{Manning14corenlp}
Christopher~D. Manning, Mihai Surdeanu, John Bauer, Jenny~Rose Finkel, Steven
  Bethard, and David McClosky. 2014.
\newblock The stanford corenlp natural language processing toolkit.
\newblock In \emph{Proceedings of the 52nd Annual Meeting of the Association
  for Computational Linguistics, System Demonstrations}.

\bibitem[{Ribeiro et~al.(2021)Ribeiro, Schmitt, Sch{\"{u}}tze, and
  Gurevych}]{Ribeiro21amrenc}
Leonardo F.~R. Ribeiro, Martin Schmitt, Hinrich Sch{\"{u}}tze, and Iryna
  Gurevych. 2021.
\newblock Investigating pretrained language models for graph-to-text
  generation.
\newblock In \emph{Proceedings of the 3rd Workshop on Natural Language
  Processing for Conversational AI}.

\bibitem[{Roy and Grangier(2019)}]{Roy19latent1}
Aurko Roy and David Grangier. 2019.
\newblock Unsupervised paraphrasing without translation.
\newblock In \emph{Proceedings of the 57th Conference of the Association for
  Computational Linguistics (ACL)}.

\bibitem[{Sachan and Xing(2016)}]{Sachan16amrsum}
Mrinmaya Sachan and Eric~P. Xing. 2016.
\newblock Machine comprehension using rich semantic representations.
\newblock In \emph{Proceedings of the 54th Annual Meeting of the Association
  for Computational Linguistics (ACL)}.

\bibitem[{Sun et~al.(2021)Sun, Ma, and Peng}]{Sun21aesop}
Jiao Sun, Xuezhe Ma, and Nanyun Peng. 2021.
\newblock {AESOP:} paraphrase generation with adaptive syntactic control.
\newblock In \emph{Proceedings of the 2021 Conference on Empirical Methods in
  Natural Language Processing (EMNLP)}.

\bibitem[{Tian et~al.(2021)Tian, Sridhar, and Peng}]{Tian21creativegen}
Yufei Tian, Arvind~Krishna Sridhar, and Nanyun Peng. 2021.
\newblock Hypogen: Hyperbole generation with commonsense and counterfactual
  knowledge.
\newblock In \emph{Findings of the Association for Computational Linguistics:
  (EMNLP-Findings)}.

\bibitem[{Wang et~al.(2019)Wang, Singh, Michael, Hill, Levy, and
  Bowman}]{Wang19glue}
Alex Wang, Amanpreet Singh, Julian Michael, Felix Hill, Omer Levy, and
  Samuel~R. Bowman. 2019.
\newblock {GLUE:} {A} multi-task benchmark and analysis platform for natural
  language understanding.
\newblock In \emph{7th International Conference on Learning Representations
  (ICLR)}.

\bibitem[{Wieting and Gimpel(2018)}]{Gimpel18paranmt}
John Wieting and Kevin Gimpel. 2018.
\newblock Paranmt-50m: Pushing the limits of paraphrastic sentence embeddings
  with millions of machine translations.
\newblock In \emph{Proceedings of the 56th Annual Meeting of the Association
  for Computational Linguistics (ACL)}.

\bibitem[{Zhang et~al.(2019)Zhang, Yang, Yuan, Shen, and Carin}]{Zhang19sivae}
Xinyuan Zhang, Yi~Yang, Siyang Yuan, Dinghan Shen, and Lawrence Carin. 2019.
\newblock Syntax-infused variational autoencoder for text generation.
\newblock In \emph{Proceedings of the 57th Conference of the Association for
  Computational Linguistics (ACL)}.

\bibitem[{Zhang et~al.(2021)Zhang, Parulian, Ji, Elsayed, Myers, and
  Palmer}]{Zhang20amrie}
Zixuan Zhang, Nikolaus~Nova Parulian, Heng Ji, Ahmed Elsayed, Skatje Myers, and
  Martha Palmer. 2021.
\newblock Fine-grained information extraction from biomedical literature based
  on knowledge-enriched abstract meaning representation.
\newblock In \emph{Proceedings of the 59th Annual Meeting of the Association
  for Computational Linguistics and the 11th International Joint Conference on
  Natural Language Processing (ACL/IJCNLP)}.

\bibitem[{Zhou et~al.(2020)Zhou, Zhang, Ji, and Tang}]{Zhou20amrparser3}
Qiji Zhou, Yue Zhang, Donghong Ji, and Hao Tang. 2020.
\newblock {AMR} parsing with latent structural information.
\newblock In \emph{Proceedings of the 58th Annual Meeting of the Association
  for Computational Linguistics (ACL)}.

\end{thebibliography}

\clearpage
\appendix
\section{Implementation Details}

We use around 20 millions of examples\footnote{\url{https://github.com/uclanlp/synpg}} in ParaNMT \cite{Gimpel18paranmt} to train AMRPG and all baselines.
The semantic encoder and the syntactic decoder are trained from scratch, with the default architecture and the default parameters of \texttt{torch.nn.Transformer}.
The max length for input sentences, the linearized constituency parses, and the linearized AMR graph are set to 40, 160, and 250, respectively.
The word dropout rate is 0.4 while the node masking rate is 0.6.
We consider Adam optimizer with the learning rate being $10^{-4}$ and the weight decay being $10^{-5}$.
The total number of epochs is set to 10.
When generating the outputs, we use random sampling with temperature being 0.5.
The model is trained with 4 NVIDIA V100 GPUs with 16 GB memory each.
It takes around 7 days to finish the training process.

\section{Experimental Settings of Syntactically Controlled Paraphrase Generation}

\subsection{Datasets}
Following previous work \cite{Huang21synpg}, our test data is: (1) 6,400 examples of ParaNMT \cite{Gimpel18paranmt}, (2) 6,400 examples of Quora \cite{Iyer17quora}, (3) 2,048 examples of PAN \cite{Madnani12pan}, and (4) 1,920 examples of MRPC \cite{Dolan04mrpc}.

\subsection{Evaluation}
\label{app:eval}

Following previous work \cite{Huang21synpg}, we consider paraphrase pairs to evaluate the performance.
Given a paraphrase pairs $(s_1, s_2)$, we use the Standford CoreNLP constituency parser \cite{Manning14corenlp} to get their parses $(p_1, p_2)$.
The input of all baselines would be $(s_1, p_2)$ and the ground truth would be $s_2$.

Assuming the generated paraphrase is $g$, 
We use BLEU score to measure the similarity between the generated paraphrase $g$ and the ground truth $s_2$.
We also calculate the template matching accuracy (TMA) by computing the exact matching accuracy of the top-2 levels of $p_g$ and $p_2$ ($p_g$ is the constituency parse of $g$). 

\section{Experimental Settings of Model Robustness}

\subsection{Training Details}
\label{app:robtrain}

We use the pre-trained SynPG parse generator to generate the full parse for each instance with the following parse templates: ``\texttt{(S(NP)(VP)(.))}'', ``\texttt{(S(VP)(.))}'', ``\texttt{(NP(NP)(.))}'', and ``\texttt{(FRAG(SBAR)(.))}''.
Then, we use the generated full parses as the parse specifications to generate paraphrases for data augmentation.
When training classifiers with data augmentation, the original instances have four times of weights as the augmented instances when computing the loss.
We use the scripts from Huggingface\footnote{\url{https://github.com/huggingface/transformers/blob/main/examples/pytorch/text-classification/run_glue.py}} with default values to train the classifiers.

\subsection{Generating Adversarial Examples}
\label{app:adv}

We use the official script\footnote{\url{https://github.com/miyyer/scpn}} of SCPN \cite{Iyyer18scpn} to generate syntactically adversarial examples.
Specifically, we consider the first five parse templates for RTE and SST-2 and first three parse templates for MRPC to generate the adversarial examples.
As long as one of the adversarial examples makes the classifier change the prediction, we count it as a successful attack on this instance.

\end{document}